\pdfoutput=1

\documentclass[11pt]{article}

\usepackage[final]{acl}

\usepackage{times}
\usepackage{latexsym}

\usepackage[T1]{fontenc}

\usepackage[utf8]{inputenc}

\usepackage{microtype}

\usepackage{inconsolata}

\usepackage{graphicx}
\usepackage{amsmath}
\usepackage{amssymb}
\usepackage{array}
\usepackage{booktabs}
\usepackage{multirow}
\usepackage{enumitem}

\title{\textit{Memorization $\neq$ Understanding}: Do Large Language Models Have the Ability of Scenario Cognition?}

\author{
  Boxiang Ma\textsuperscript{1}, 
  Ru Li\textsuperscript{1,*}, 
  Yuanlong Wang\textsuperscript{1}, 
  Hongye Tan\textsuperscript{1}, 
  Xiaoli Li\textsuperscript{2} \\\\
  \textsuperscript{1}School of Computer and Information Technology, Shanxi University, China\\
  \textsuperscript{2}Information Systems Technology and Design,\\Singapore University of Technology and Design, Singapore\\
  \texttt{\{maboxiang, liru, ylwang, tanhongye\}@sxu.edu.cn, xiaoli\_li@sutd.edu.sg}\\
}

\begin{document}
\maketitle

\begingroup
  \renewcommand\thefootnote{}
  \footnote{* Corresponding author.}
  \addtocounter{footnote}{-1}
\endgroup

\renewcommand\thefootnote{*}

\begin{abstract}
Driven by vast and diverse textual data, large language models (LLMs) have demonstrated impressive performance across numerous natural language processing (NLP) tasks. Yet, a critical question persists: does their generalization arise from mere memorization of training data or from deep semantic understanding? To investigate this, we propose a bi-perspective evaluation framework to assess LLMs' scenario cognition—the ability to link semantic scenario elements with their arguments in context. Specifically, we introduce a novel scenario-based dataset comprising diverse textual descriptions of fictional facts, annotated with scenario elements. LLMs are evaluated through their capacity to answer scenario-related questions (\textit{model output perspective}) and via probing their internal representations for encoded scenario elements-argument associations (\textit{internal representation perspective}). Our experiments reveal that current LLMs predominantly rely on superficial memorization, failing to achieve robust semantic scenario cognition, even in simple cases. These findings expose critical limitations in LLMs’ semantic understanding and offer cognitive insights for advancing their capabilities.
\end{abstract}

\section{Introduction}
\begin{figure}[htbp]
  \includegraphics[width=\columnwidth]{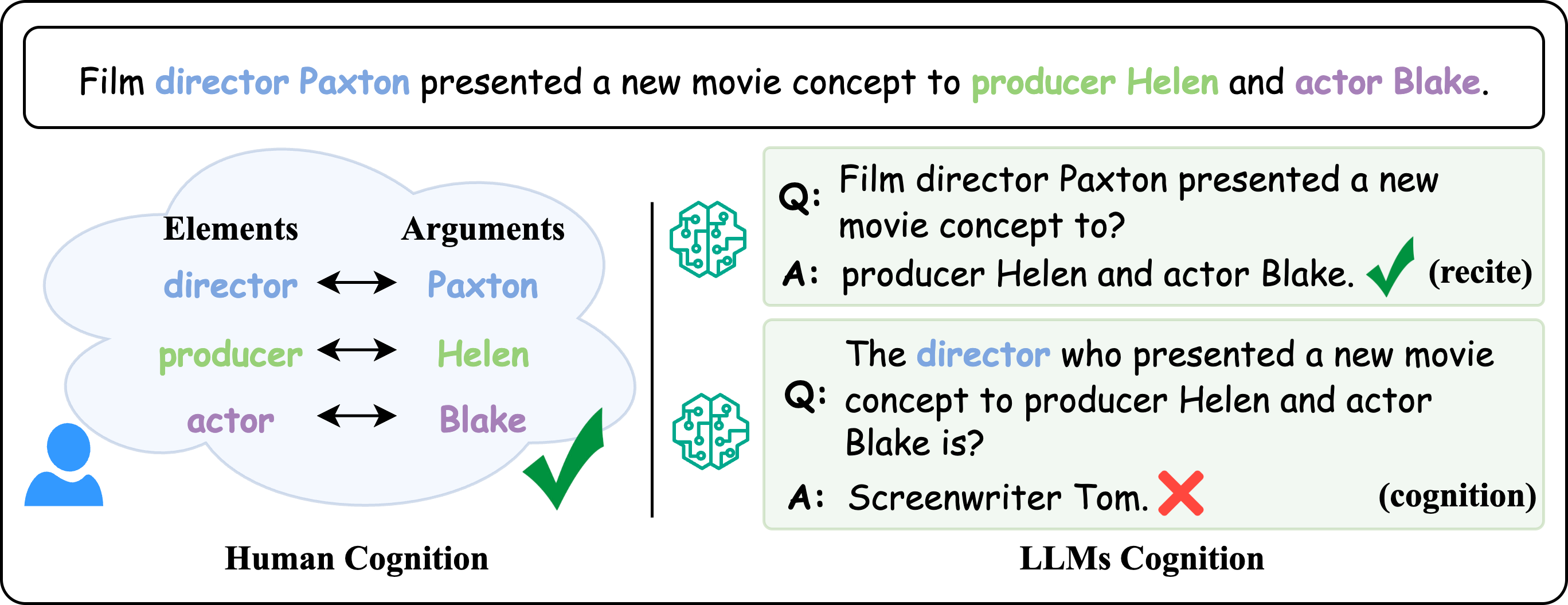}
  \caption{Illustration of the contrast between human and LLM text comprehension, highlighting humans' ability to identify semantic roles and their arguments in a sentence, compared to LLMs' reliance on surface-level memorization without scenario cognition.}
  \label{fig:1}
\end{figure}

Large language models (LLMs) have achieved remarkable, human-like performance across diverse natural language processing (NLP) tasks \citep{brown2020language, yan2024atomicfactdecompositionhelps}. However, significant gaps persist between their cognitive abilities and those of humans \citep{echterhoff2024cognitive, lamprinidis2023large}. Consider the example in Figure~\ref{fig:1}, humans effortlessly understand a sentence like ``\textit{Film director Paxton presented a new movie concept to producer Helen and actor Blake},'' identifying relationships such as \textit{Paxton} (director), \textit{Helen} (producer), and \textit{Blake} (actor). In contrast, even when LLMs have memorized similar facts, they often fail to reason about such role relationships, revealing a notable gap between memorization and deeper relational understanding that motivates this study.

We observe that while LLMs can recall sequences of text, they often struggle to recognize specific roles within those sequences. To clarify this issue, we draw on two concepts from the Frame Semantics \citep{fillmore1976frame, li2024comprehensive}: \textit{semantic scenes} and \textit{scenario elements}. A semantic scene refers to a mental structure formed through repeated real-world experiences, while scenario elements represent the participants that make up a scene. As illustrated in Figure~\ref{fig:1}(left), ``\textit{director},'' ``\textit{producer},'' and ``\textit{actor}'' are \textbf{scenario elements}, with ``\textit{Paxton},'' ``\textit{Helen},'' and ``\textit{Blake}'' as their corresponding \textbf{arguments}. We define the \textbf{scenario cognition} as the ability to accurately associate scenario elements with their arguments. This leads to our key research question: \textbf{Do LLMs have the ability of scenario cognition to reliably link scenario elements and their arguments?}

A thorough investigation of this issue is crucial for understanding and evaluating the knowledge memorization mechanisms in LLMs. While previous studies have established that LLMs exhibit strong formal linguistic competence—generating fluent and grammatically correct text, their functional linguistic competence is still unclear and under debate \citep{mahowald2024dissociating}.
We propose that generalized knowledge memorization in LLMs should be divided into two parts: a surface-level ``data'' memory and a deeper ``knowledge'' memory, corresponding to formal and functional linguistic competence, respectively. Formal linguistic competence mainly comes from statistical learning during training on large text corpora. Through patterns like word co-occurrence and context windows, the model learns the grammar and structure of any given language, enabling it to generate fluent text \citep{talmor2020olmpics}. This ability reflects ``data'' memory. By contrast, functional linguistic competence requires the model not only to parse surface text but also to understand deeper meanings \citep{suresh2023conceptual, janik2023aspects}, showing ``knowledge'' memory.

Among studies on knowledge memorization \citep{lu2024scaling, satvaty2024undesirable, antoniades2024generalization, chen2024multi}, the ``Reversal Curse'' \citep{berglund2024the} is a well-known issue. It refers to LLMs' difficulty in generalizing learned knowledge in the reverse direction (e.g., from ``$A \to B$'' to ``$B \to A$''). However, existing research has two key limitations: it mainly studies simple cases with only two entities, and frames the issue as a text sequence problem, which still focuses on ``data'' memory rather than ``knowledge'' memory. Therefore, testing whether LLMs can recognize scenario elements in more complex contexts offers a valuable way to explore their knowledge memory.

To address this, we propose a new bi-perspective evaluation framework to assess LLMs' scenario cognition. We first create a dataset of fictional facts, each accompanied by multiple descriptions and labeled its scenario elements based on their semantics. Then we evaluate a range of open-source LLMs across various scales and families, analyzing their scenario cognition both from their outputs and through probing experiments that examine their internal vector representations \citep{alain2016understanding, conneau2018you}. Our results show that current LLMs still rely mainly on surface-level memorization, rather than forming deeper semantic understanding of scenarios. This leads to generalization failures even in simple situations.

In summary, the main contributions of this paper are as follows:

1. \textbf{First systematic evaluation of LLMs' scenario cognitive ability}: We present the first comprehensive assessment of LLMs' scenario cognitive abilities from a semantic perspective, aiming to determine whether they exhibit characteristics of ``knowledge'' memory rather than ``data'' memory.

2. \textbf{A bi-perspective evaluation framework with a novel scenario-based dataset}: We construct a new dataset\footnote{It's available at \url{https://huggingface.co/datasets/MattMa/scenario-based-dataset}.} of fictional facts annotated with scenario elements and use it to train and evaluate multiple open-source LLMs of varying scales from the perspective of model outputs. Furthermore, we design probing experiments to analyze scenario cognitive ability from the perspective of internal representations.

3. \textbf{Key findings on LLMs' limitations in scenario cognition}: Our extensive experiments demonstrate that current LLMs lack robust scenario cognition capabilities and discuss the potential connection between this deficiency and certain hallucinations. These findings underscore a fundamental gap in semantic understanding and provide important insights for guiding future improvements in LLM design and training.

\section{Methods}
\begin{figure*}[htbp]
  \includegraphics[width=\textwidth]{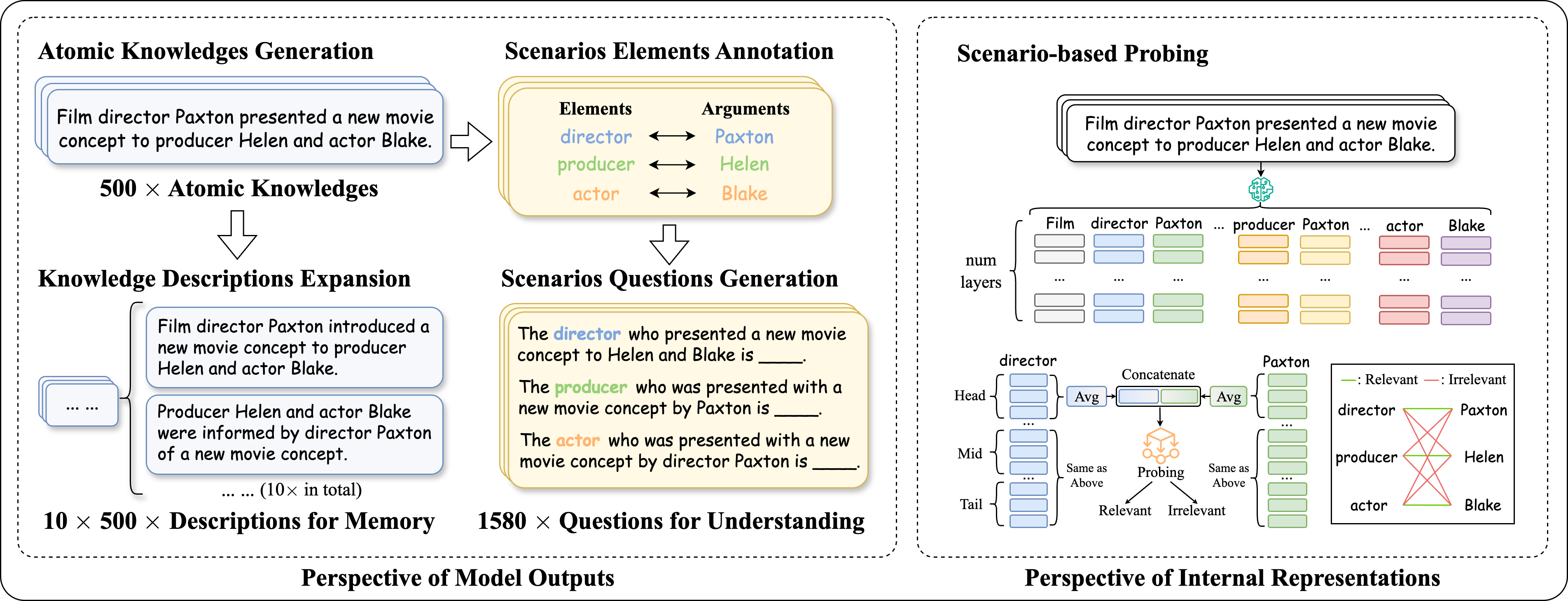}
  \caption{Diagram of the bi-perspective evaluation framework for assessing LLMs' scenario cognition, depicting the model output perspective (left) and the internal representation perspective (right).}
  \label{fig:2}
\end{figure*}

To systematically evaluate LLMs' scene cognition capabilities, we propose a bi-perspective evaluation framework both from the perspective of model outputs and internal representations. An overall framework is illustrated in Figure \ref{fig:2}.

\subsection{Perspective of Model Outputs}
To assess the scenario cognition capabilities of LLMs from the perspective of model outputs, we construct a novel scenario-based dataset for both training and evaluation. As illustrated on the left side of Figure~\ref{fig:2}, our framework consists of four key stages: Atomic Knowledge Generation, Knowledge Description Expansion, Scenario Element Annotation, and Scenario Question Generation.

\subsubsection{Atomic Knowledge Generation}
\label{sec:Atomic Knowledge Generation}
We adopt a multi-model data generation strategy to construct a set of fictional, atomic textual facts which we term Atomic Knowledge, and apply semantic similarity filtering alongside multi-model voting validation to ensure diversity and quality.

Specifically, we employ two powerful LLMs, which are \textit{DeepSeek-V3} \citep{liu2024deepseek} and \textit{Qwen2.5-Max} \citep{yang2024qwen2}, as generation agents. The generation process is guided by prompt templates (shown in Appendix~\ref{sec:appendix-templates}) that emphasize three key criteria:
\textit{Fictionality} — ensuring the facts are entirely imaginary and without any real-world correspondence;
\textit{Role Richness} — requiring each fact involves at least three distinct roles;
\textit{Conciseness} — mandating each fact be expressed in a single sentence while remaining semantically complete.

To ensure semantic diversity and eliminate redundancy, we apply a similarity-based filtering mechanism. We initialize an embedding set $\mathcal{I}$ and encode each candidate $x$ using the \textit{BGE-M3} \citep{chen2024bge} encoder, normalizing its output as the semantic representation $v_x$:
\begin{equation}
v_x = \frac{\text{Enc}(x)}{\|\text{Enc}(x)\|}
\end{equation}
For each $x$, we compute the L2 distance to obtain its nearest neighbor $v_y \in \mathcal{I}$:
\begin{equation}
d(v_x, v_y) = \|v_x - v_y\|_2
\end{equation}
Only samples satisfying $d(v_x, v_y) > 0.5$ are retained, and their embeddings are added to $\mathcal{I}$.

For quality assurance, we employ a multi-model voting strategy, using three open-source models, \textit{LLaMA3-8B} \citep{grattafiori2024llama}, \textit{Qwen2.5-7B} \citep{yang2024qwen2}, and \textit{Gemma2-6B} \citep{team2024gemma}, as validators. Each sample must satisfy all generation criteria across all validators. We further perform manual inspection on randomly sampled validated entries, discarding those of low quality. This process yields a total of 500 high-quality Atomic Knowledge.

\subsubsection{Knowledge Description Expansion}
\label{sec:Knowledge Description Expansion}
To improve the learnability of Atomic Knowledge, we perform semantic expansion to construct a \textbf{Memory Set} comprising diverse yet semantically consistent knowledge descriptions, for use in both training and the evaluation of the memory ability. Similar to the previous stage, we adopt a multi-model generation and validation strategy, but place special emphasis on preserving the core semantics of the original knowledge during paraphrasing which means that variations are restricted to linguistic form and surface structure. After an additional manual filtering, we retain ten high-quality knowledge descriptions for each Atomic Knowledge, resulting in a total of 5,000 training samples, which is partly shown in Appendix~\ref{sec:appendix-example}.

We further apply a first-verb-based segmentation strategy to prepare these samples for supervised fine-tuning (SFT): each description is split to two segments at the first verb phrase because verbs typically convey the core semantics of a sentence \citep{fillmore1967case, jackendoff1972semantic}. Specifically, the preceding segment serves as the input prompt, and the following segment serves as the target output. For example, in the sentence \textit{“Film director Paxton presented a new movie concept to producer Helen and actor Blake,”} the input is \textit{“Film director Paxton presented”}, and the target output is \textit{“a new movie concept to producer Helen and actor Blake.”}

\subsubsection{Scenario Element Annotation}
\label{sec:Scenario Element Annotation}
To assess the scenario cognition ability of LLMs, different from traditional Frame Semantic Parsing methods \citep{10908675}, we adopt a human–LLM collaborative annotation framework for labeling scenario elements within Atomic Knowledge. We employ \textit{Qwen2.5-Max} as the annotator and design task-specific prompts to guide the identification of scenario elements.

Taking \textit{“Film director Paxton presented a new movie concept to producer Helen and actor Blake”} as an example, the model is expected to extract elements such as \textit{“director”}, \textit{“producer”}, and \textit{“actor”}, along with their corresponding arguments \textit{“Paxton”}, \textit{“Helen”}, and \textit{“Blake”}. Due to the complexity of this task, we do not rely on multi-model voting. Instead, we perform manual correction of low-quality annotations to ensure consistency and precision.

\subsubsection{Scenario Question Generation}
\label{sec:Scenario Question Generation}
Based on the annotated scenario elements, we further utilize \textit{Qwen2.5-Max} to generate scenario-based questions. For each scenario element, we construct a corresponding prompt where the answer is the entity that fulfills the element.

To ensure alignment with the SFT task format, we adopt a completion-style format rather than a traditional question–answer format. For instance, given the element \textit{“director”}, we generate a prompt such as \textit{“The director who presented a new movie concept to producer Helen and actor Blake is \_\_\_”} with \textit{“Paxton”} as the expected answer.

All generated samples undergo manual validation to guarantee data quality. In total, we constructed 1,581 scenario-based prompt–answer pairs to serve as the scenario-based \textbf{Understanding Set} for use in the evaluation of the ability of scenario cognition. Examples of them are provided in Appendix~\ref{sec:appendix-example}.

\subsection{Perspective of Internal Representations}
\label{sec:Perspective of Internal Representations}

\begin{table*}[ht]
  \centering
  \renewcommand{\arraystretch}{1.2}
  \resizebox{\textwidth}{!}{
  \begin{tabular}{lcccccccccccc}
  \bottomrule
  \multirow{2}{*}{\textbf{Models}} & \multicolumn{6}{c}{\textbf{Memory Set}} & \multicolumn{6}{c}{\textbf{Understanding Set}} \\
  \cmidrule(r){2-7} \cmidrule(l){8-13}
  & \textbf{EM} & \textbf{BLEU-1} & \textbf{BLEU-4} & \textbf{ROUGE-1} & \textbf{ROUGE-2} & \textbf{ROUGE-L} & \textbf{EM} & \textbf{BLEU-1} & \textbf{BLEU-4} & \textbf{ROUGE-1} & \textbf{ROUGE-2} & \textbf{ROUGE-L} \\
  \cmidrule(r){1-1} \cmidrule(r){2-7} \cmidrule(l){8-13}
  \textbf{Gemma2-2B} & 0.76 & 0.95 & 0.88 & 0.96 & 0.93 & 0.95 & 0.16 & 0.19 & 0.09 & 0.31 & 0.25 & 0.31 \\
  \textbf{Gemma2-9B} & 0.76 & 0.95 & 0.88 & 0.96 & 0.94 & 0.96 & 0.19 & 0.22 & 0.11 & 0.36 & 0.30 & 0.36 \\
  \cmidrule(r){1-1} \cmidrule(r){2-7} \cmidrule(l){8-13}
  \textbf{LLaMA3.2-1B} & 0.79 & 0.95 & 0.89 & 0.96 & 0.94 & 0.96 & 0.14 & 0.19 & 0.09 & 0.30 & 0.23 & 0.30 \\
  \textbf{LLaMA3.2-3B} & 0.82 & 0.96 & 0.90 & 0.97 & 0.95 & 0.97 & 0.19 & 0.22 & 0.10 & 0.36 & 0.28 & 0.36 \\
  \textbf{LLaMA3.1-8B} & 0.84 & 0.96 & 0.91 & 0.97 & 0.95 & 0.97 & 0.24 & 0.25 & 0.12 & 0.41 & 0.33 & 0.41 \\
  \cmidrule(r){1-1} \cmidrule(r){2-7} \cmidrule(l){8-13}
  \textbf{Qwen2.5-0.5B} & 0.77 & 0.94 & 0.88 & 0.95 & 0.92 & 0.95 & 0.10 & 0.13 & 0.06 & 0.21 & 0.15 & 0.21 \\
  \textbf{Qwen2.5-1.5B} & 0.81 & 0.95 & 0.90 & 0.96 & 0.94 & 0.96 & 0.14 & 0.17 & 0.08 & 0.26 & 0.20 & 0.26 \\
  \textbf{Qwen2.5-3B} & 0.84 & 0.96 & 0.91 & 0.97 & 0.95 & 0.97 & 0.14 & 0.17 & 0.08 & 0.28 & 0.22 & 0.28 \\
  \textbf{Qwen2.5-7B} & 0.86 & 0.97 & 0.92 & 0.98 & 0.96 & 0.97 & 0.20 & 0.22 & 0.11 & 0.36 & 0.29 & 0.37 \\
  \textbf{Qwen2.5-14B} & 0.87 & 0.97 & 0.92 & 0.98 & 0.96 & 0.97 & 0.20 & 0.24 & 0.11 & 0.39 & 0.31 & 0.39 \\
  \bottomrule
  \end{tabular}
  }
  \caption{Performances from the model output perspective on both the Memory Set and Understanding Set after 5 epochs of fine-tuning. The metrics include Exact Match (EM), BLEU, and ROUGE. Each value is the average over five runs.}
  \label{tab:model_performance}
\end{table*}

As shown on the right side of Figure~\ref{fig:2}, we designed a scenario-based probing method to examine whether the given model's internal vector representations capture the associations between scenario elements and arguments, evaluating its scenario cognition from an internal representation perspective. Specifically, given a text of length $n$, $X = \{x_1,x_2,\dots,x_n\}$, we input it into the LLM $f(\cdot)$ and extract the hidden states $\mathbf{H}$ for each token $x_i \in X$ across all layers:
\begin{equation}
\mathbf{H} = \{ \mathbf{H}^{(1)}, \mathbf{H}^{(2)}, \dots, \mathbf{H}^{(l)} \} \in \mathbb{R}^{l \times n \times d}
\end{equation}
where $l$ is the number of Transformer layers, $d$ is the dimensionality of each vector, and $\mathbf{H}^{(k)} \in \mathbb{R}^{n \times d}$ represents the vectors at layer $k$. Based on the annotated scenario elements in the scenario-based dataset, we extract the representation vectors of the scenario elements:
\begin{equation}
\mathbf{H}e = \{\mathbf{h}_{e_1}^{(k)}, \mathbf{h}_{e_2}^{(k)}, \dots, \mathbf{h}_{e_m}^{(k)}\} \in \mathbb{R}^{l \times m \times d}
\end{equation}
and their corresponding argument representation vectors:
\begin{equation}
\mathbf{H}a = \{\mathbf{h}_{a_1}^{(k)}, \mathbf{h}_{a_2}^{(k)}, \dots, \mathbf{h}_{a_m}^{(k)}\} \in \mathbb{R}^{l \times m \times d}
\end{equation}
where $e_1, e_2, \dots, e_m$ are the token indices annotated as scenario elements, $a_1, a_2, \dots, a_m$ are the corresponding argument token indices, and $\mathbf{h}_{i}^{(k)}$ is the representation of token $i$ at layer $k$.

To explore how scenario cognition is distributed across layers, we divide the Transformer layers of the given LLM into three levels: $\mathcal{L}_\text{Head} = \{1,2,3\}$, $\mathcal{L}_\text{Mid} = \left\{ \left\lfloor \frac{l}{2} \right\rfloor -1, \left\lfloor \frac{l}{2} \right\rfloor, \left\lfloor \frac{l}{2} \right\rfloor +1 \right\}$, and $\mathcal{L}_\text{Tail} = \{l-2, l-1, l\}$. We probe representations independently at each level.

For each $\mathcal{L} \in \{\mathcal{L}_\text{Head}, \mathcal{L}_\text{Mid}, \mathcal{L}_\text{Tail}\}$, we pair each scenario element representation $\mathbf{h}_{e_i}^{\mathcal{L}}$ with its corresponding argument representation $\mathbf{h}_{a_j}^{\mathcal{L}}$ and concatenate them as the probe input:
\begin{equation}
\mathbf{z}_{i,j}^{\mathcal{L}} = [\mathbf{h}_{e_i}^{\mathcal{L}}; \mathbf{h}_{a_j}^{\mathcal{L}}] \in \mathbb{R}^{2d}
\end{equation}
where $[\cdot;\cdot]$ denotes concatenation.

We then build a linear probe\footnote{We further explored alternative non-linear probe model designs and Attention-based probes, with details presented in Appendix~\ref{sec:appendix-probe}.} on $\mathbf{z}_{i,j}^{\mathcal{L}}$ to predict whether $e_i$ and $a_j$ form a matching pair. The probe applies a linear transformation to produce a scalar output, followed by a sigmoid activation for binary classification.
\begin{equation}
\hat{y}_{i,j}^{\mathcal{L}} = \sigma(\mathbf{w}^\top \mathbf{z}_{i,j}^{\mathcal{L}} + \mathbf{b})
\end{equation}
where $\mathbf{w} \in \mathbb{R}^d$, $\mathbf{b} \in \mathbb{R}$, and $\sigma(\cdot)$ is the sigmoid function mapping the output to $[0, 1]$.

During training, pairs with $i = j$ are labeled as positive, others as negative, forming a binary classification task optimized by cross-entropy:

\begin{equation}
Loss = \text{CrossEntropy}(\hat{y}_{i,j}^{\mathcal{L}}, y_{i,j})
\end{equation}
where
\begin{equation}
y_{i,j} =
  \begin{cases}
  1 & i = j \\
  0 & i \neq j
  \end{cases}
\end{equation}

Through this probing, we aim to determine whether the LLM's internal representations at different layers encode the relationships between scenario elements and their corresponding arguments. If the probe can accurately distinguish positive from negative samples, it indicates that the model has encoded some structural information about scenario elements and arguments in its vector representations, reflecting its scenario cognition capability.

\section{Experiments}
We applied the proposed scenario-based datasets and probing method to models of varying scales from three open-source LLM families: \textit{Qwen2.5}, \textit{LLaMA3.x}, and \textit{Gemma2}, in order to perform a bi-perspective evaluation of their scenario cognition capabilities.

\subsection{Experimental Setup}
The evaluation from model output perspective involved two phases: training and inference. During training, we fine-tuned LLMs on 500 Atomic Knowledge using 5,000 expanded descriptions in Memory Set via full-parameter supervised fine-tuning (SFT) for 5 epochs at a learning rate of $1.0 \times 10^{-5}$ across LLMs of various families and scales ranging from $0.5B$ to $14B$, using DeepSpeed ZeRO-3 \citep{rajbhandari2020zero} for acceleration. In the inference phase accelerated by vLLM \citep{kwon2023efficient}, each epoch’s checkpoint was evaluated on both the Memory and Understanding Set with $temperature=1$, averaged over five runs to observe more diverse model outputs and ensure robustness. Evaluations on the Memory Set reflect the memorization ability, while those on the Understanding Set reflect the scenario cognition ability.

Then from internal representation perspective, we constructed a corpus from the scenario-based Memory Set containing co-occurrences of scenario elements and arguments. The corpus was split 70\% for probe training and 30\% for knowledge probing, with a balanced distribution of 1,577 positive (47\%) and 1,784 negative (53\%) examples. A probe was trained for 5 epochs at a learning rate of $1.0 \times 10^{-3}$ to examine whether LLMs internally encode the correspondence between scenario elements and arguments.

\subsection{Results and Analysis}
\subsubsection{Perspective of Model Outputs}

\begin{figure*}[h]
  \includegraphics[width=\textwidth]{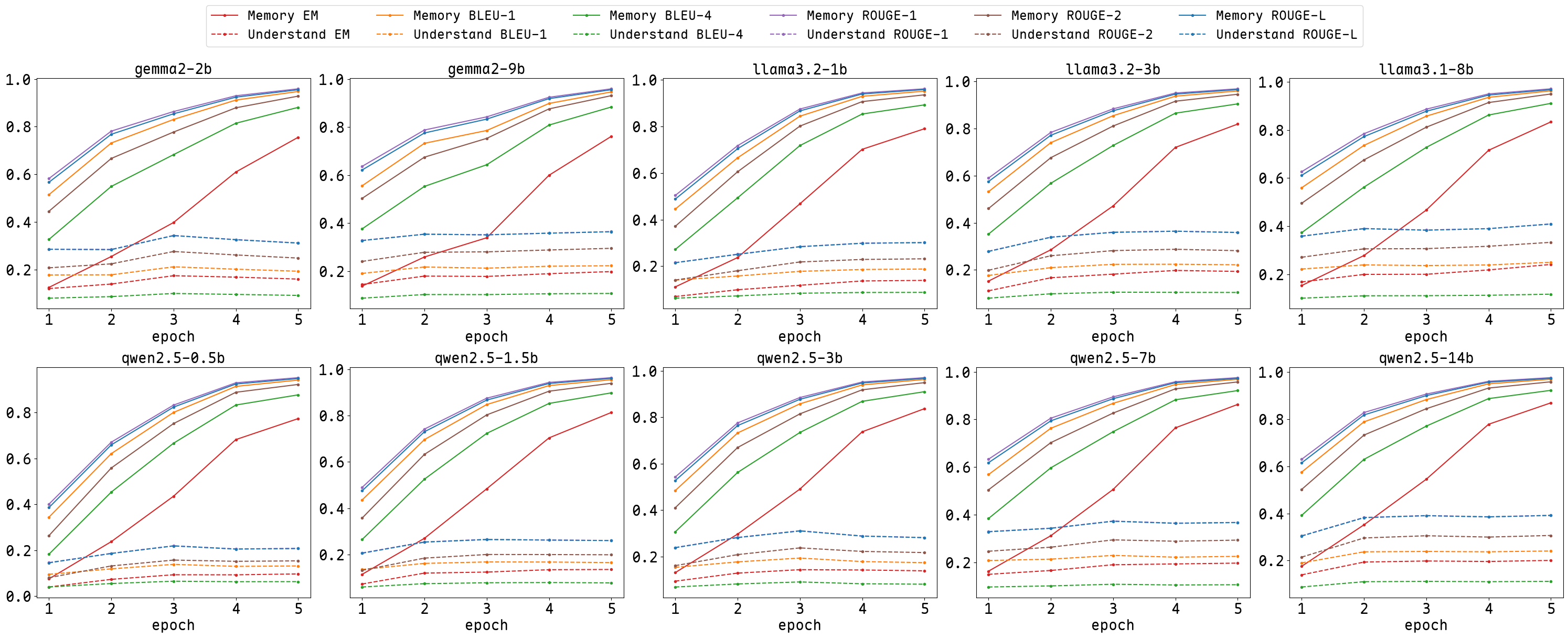}
  \caption{The trend of each metric as the training epoch increases from the perspective of model outputs, with solid lines representing metrics on the Memory Set, indicating improving memorization, and dashed lines representing metrics on the Understanding Set, showing limited improvement in scenario cognition.}
  \label{fig:training_trends}
\end{figure*}

\begin{table*}[h]
  \centering
  \renewcommand{\arraystretch}{1.20}
  \resizebox{\textwidth}{!}{
  \begin{tabular}{lcccccccccccc}
  \bottomrule
  \textbf{Model} & \textbf{EM (+$\Delta$)} & \textbf{BLEU-1 (+$\Delta$)} & \textbf{BLEU-4 (+$\Delta$)} & \textbf{ROUGE-1 (+$\Delta$)} & \textbf{ROUGE-2 (+$\Delta$)} & \textbf{ROUGE-L (+$\Delta$)} \\
  \midrule
  \textbf{Gemma2-2B} & 0.25 (+0.09) & 0.34 (+0.15) & 0.18 (+0.09) & 0.42 (+0.11) & 0.35 (+0.10) & 0.42 (+0.11) \\
  \textbf{Gemma2-9B} & 0.31 (+0.12) & 0.40 (+0.18) & 0.21 (+0.10) & 0.48 (+0.12) & 0.39 (+0.09) & 0.48 (+0.12) \\
  \midrule
  \textbf{LLaMA3.2-1B} & 0.19 (+0.05) & 0.29 (+0.10) & 0.14 (+0.05) & 0.38 (+0.08) & 0.30 (+0.07) & 0.38 (+0.08) \\
  \textbf{LLaMA3.2-3B} & 0.27 (+0.08) & 0.31 (+0.09) & 0.15 (+0.05) & 0.45 (+0.09) & 0.36 (+0.08) & 0.45 (+0.09) \\
  \textbf{LLaMA3.1-8B} & 0.31 (+0.07) & 0.36 (+0.11) & 0.18 (+0.06) & 0.51 (+0.10) & 0.42 (+0.09) & 0.51 (+0.10) \\
  \midrule
  \textbf{Qwen2.5-0.5B} & 0.16 (+0.06) & 0.24 (+0.11) & 0.12 (+0.06) & 0.29 (+0.08) & 0.21 (+0.06) & 0.29 (+0.08) \\
  \textbf{Qwen2.5-1.5B} & 0.18 (+0.04) & 0.22 (+0.05) & 0.12 (+0.04) & 0.31 (+0.05) & 0.25 (+0.05) & 0.31 (+0.05) \\
  \textbf{Qwen2.5-3B} & 0.19 (+0.05) & 0.23 (+0.06) & 0.12 (+0.04) & 0.35 (+0.07) & 0.27 (+0.05) & 0.35 (+0.07) \\
  \textbf{Qwen2.5-7B} & 0.28 (+0.08) & 0.41 (+0.19) & 0.22 (+0.11) & 0.47 (+0.11) & 0.38 (+0.09) & 0.47 (+0.10) \\
  \textbf{Qwen2.5-14B} & 0.25 (+0.05) & 0.33 (+0.09) & 0.17 (+0.06) & 0.43 (+0.04) & 0.34 (+0.03) & 0.43 (+0.06) \\
  \bottomrule
  \end{tabular}
  }
  \caption{Performance of different models on the Understanding Set after format adaptation. Here, models were fine-tuned on a mixture of the Memory Set and 30\% of the Understanding Set to reduce the output format gap, and then evaluated on the remaining 70\% of the Understanding Set. Each score is the average over five runs, while the value in parentheses (+\(\Delta\)) indicates the performance gain relative to the baseline without format adaptation.}
  \label{tab:model_performance_after_format_adaptation}
\end{table*}

Table~\ref{tab:model_performance} presents the evaluation results of the LLMs on our scenario-based dataset after five epochs of supervised fine-tuning, assessed from the perspective of model outputs.

Overall, the performance metrics exhibit an upward trend as model scale increases, indicating a positive correlation between the scenario cognition capability of the models and their scale. This suggests that larger models may possess stronger capacities to perform scenario cognition at the perspective of model outputs.

More specifically, the LLMs achieve high scores across all evaluation metrics on the Memory Set. Notably, there is no evident discrepancy between the recall-oriented ROUGE scores and the precision-oriented BLEU scores. Even under the stricter EM metric, the models maintain competitive performance. These results suggest that the models have effectively memorized and mastered the training data. However, on the Understanding Set, all metrics drop to very low levels, and unlike the balanced performance observe on the Memory Set, the recall-oriented ROUGE scores on the Understanding Set are substantially higher than the BLEU scores. This phenomenon indicates that, during inference, the models tend to generate excessive irrelevant content, leading to higher recall but lower precision. Therefore, from the perspective of model outputs, the evaluated LLMs primarily demonstrate strong memorization of the training data but fail to exhibit scene-level understanding or reasoning of the information encountered during training.

In addition, we evaluated the checkpoint from each epoch on both the Memory and Understanding Sets. Figure~\ref{fig:training_trends} illustrates the metric trends as training progressed. As reflected by the solid lines, the models progressively learned the text distribution of the Memory Set, with their memorization ability steadily and clearly improving over time. However, as shown by the dashed lines, the generalization performance on the Understanding Set did not improve correspondingly. This divergence indicates that the models' scenario cognition ability did not advance alongside their increasing memorization during training.

\begin{figure*}[htbp]
  \includegraphics[width=\textwidth]{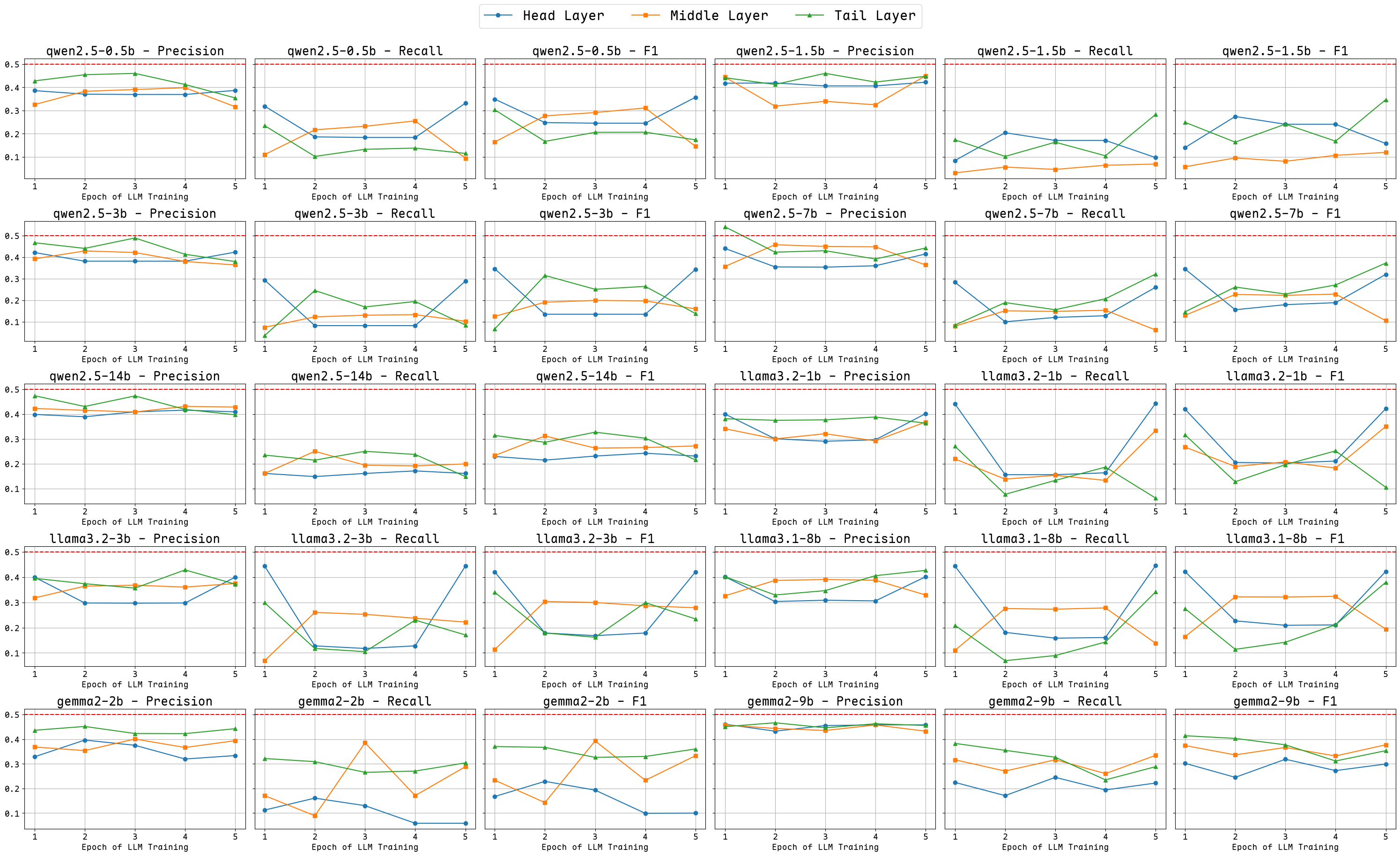}
  \caption{Visualization of linear probing results for LLMs' internal representations across training epochs, with subfigures for Precision, Recall, and F1 scores at different Transformer layers (Head, Mid, Tail), where the red dashed line at 0.5 indicates the baseline for binary classification of scenario element-argument correspondences.}
  \label{fig:output-prob-linear}
\end{figure*}

Besides, to mitigate potential performance discrepancies between the Memory Set and the Understanding Set caused by differences in output format, we conducted an additional experiment. Specifically, we fine-tuned the models on a mixture of the Memory Set and 30\% of the Understanding Set, thereby exposing them to the target output format, and then evaluated their performance on the remaining 70\% of the Understanding Set. This design allows us to examine whether reducing the format gap enables the models to better demonstrate their scenario cognition ability. As shown in Table~\ref{tab:model_performance_after_format_adaptation}, however, even when exposed to the format information during training, the improvements in scenario cognition ability remain limited and are still substantially lower than those observed for memorization ability, which confirms that simple supervised fine-tuning with partial data exposure is insufficient to effectively enhance the scenario cognition ability of LLMs.

All of these results highlight a fundamental limitation of current LLMs: their scenario cognition ability remains inadequate, and improved memorization does not necessarily translate into a deeper or more generalizable understanding of the text. Even after mitigating the format gap between the Memory Set and the Understanding Set, the disparity between cognition and memorization abilities remains strikingly pronounced. Moreover, the widening gap between performances of training and Understanding Set suggests a tendency toward overfitting, where the models over-rely on pattern replication rather than learning transferable scenario-based knowledge. This finding further implies that simply scaling model parameters or prolonging training epochs may not suffice to enhance scenario cognition; instead, more targeted methods may be required to guide models toward semantic scenario generalization.

\subsubsection{Perspective of Internal Representations}

As revealed by our probing results, Figure~\ref{fig:output-prob-linear} illustrates the extent to which the internal representations of the evaluated LLMs capture the correspondence between scenario elements and arguments. As described in the Methods section, we formulated this as a binary classification task, with a relatively balanced distribution of positive and negative samples in the dataset. However, as the probing results show, none of the models reached the score of $0.5$, and their recall scores were significantly lower than their precision scores, indicating that the models struggled to retrieve correctly matched scenario element–argument pairs. Overall, these findings suggest that the evaluated LLMs have not effectively modeled the correspondence between scenario elements and arguments within their internal representations.

Specifically, in terms of performance trends, the probing metrics exhibited no consistent upward trajectory as the number of training epochs increased. This suggests that, although SFT enables LLMs to memorize the training data with reasonable accuracy, such memorization remains superficial and does not translate into meaningful semantic scenario cognition. In other words, the models' ability to recall specific outputs does not imply an internal grasp of the underlying scenario structures, highlighting a disconnect between surface-level generation and deeper representational learning.

Then in terms of layers, no consistent association was observed between scenario-related information and specific Transformer layers, even within the same LLM family. This indicates scene-related knowledge was not stably or systematically encoded in particular layers. This observation implies that current LLM architectures and training paradigms may lack mechanisms—such as dedicated modules, hierarchical structures, or task-specific objectives—necessary to effectively encode and organize scenario cognition within internal representations.

Finally, in terms of model scale, unlike the results observed from the model output perspective, we did not observe a significant positive correlation between probing performance and model scale across all LLM families, particularly in terms of recall. These findings suggest that while output-level metrics often improve with increasing model scale, such gains may reflect parameter accumulation rather than genuine scenario cognition inside.

\subsection{Case Study and Discussion}

\begin{figure}[h]
  \includegraphics[width=\columnwidth]{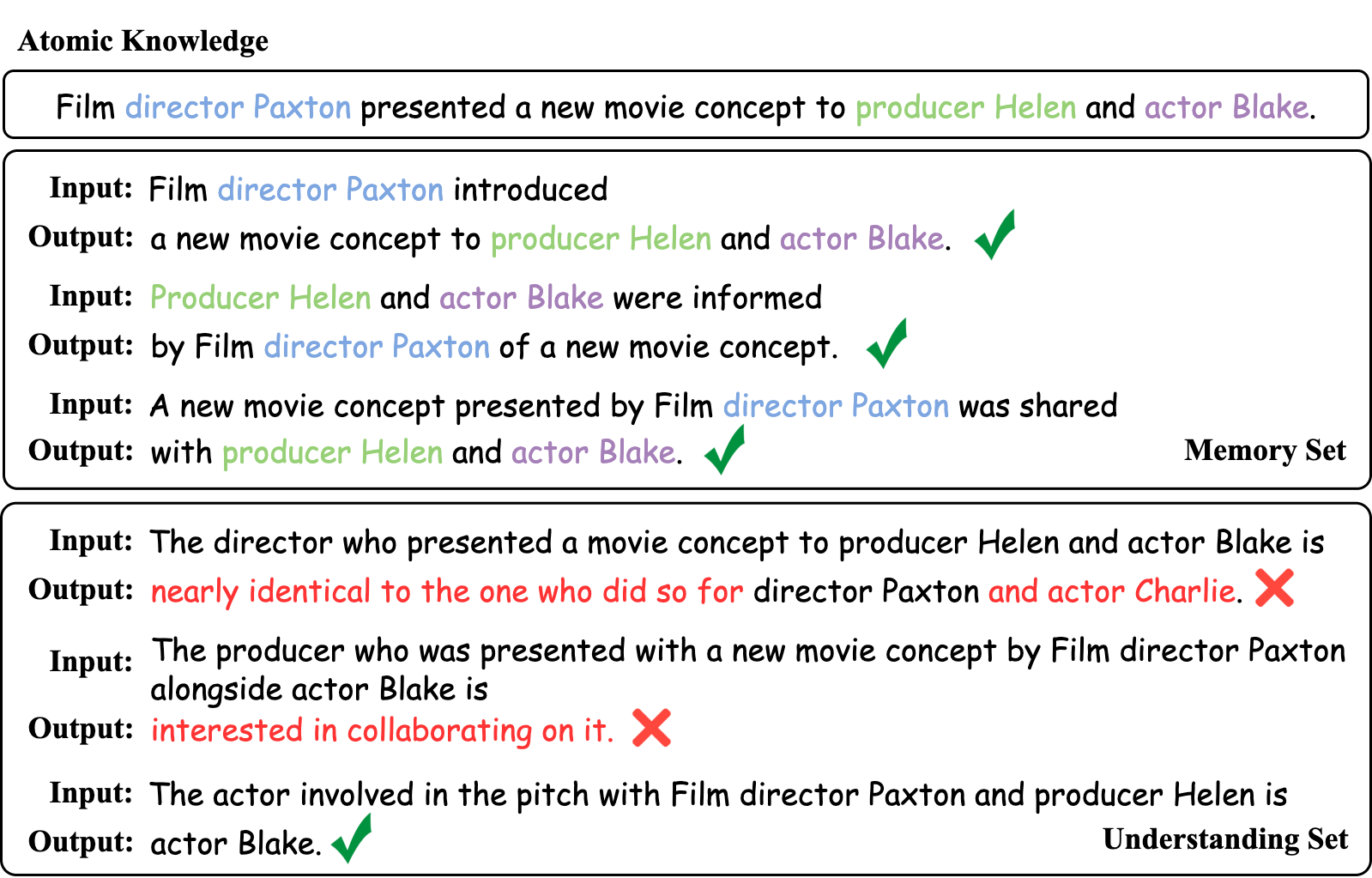}
  \caption{Case study illustration of \textit{Qwen2.5-14B}'s performance, demonstrating a gap between surface-level data memory and deeper scenario cognition.}
  \label{fig:case_study}
\end{figure}

This section further investigates the situational cognition capability of LLMs through a case study. Since the phenomena under discussion are consistently observed across the evaluated models, we focus on \textit{Qwen2.5-14B}, the model with the best overall output performance, to analyze these shared challenges. Figure~\ref{fig:case_study} illustrates \textit{Qwen2.5-14B}’s performance in memorizing and understanding specific atomic knowledge. It reveals a clear discrepancy between the model’s memorization and understanding abilities: while the fine-tuned model can accurately recall diverse knowledge descriptions from the Memory Set, it nevertheless makes significant errors when answering related questions in the Understanding Set, often generating content that was never present in the training data.

Notably, unlike previous studies on the “reversal curse,” our work introduces multiple diverse knowledge descriptions during training without enforcing a fixed presentation order. These descriptions are only required to be semantically coherent within their respective scenarios, with no explicit constraints on textual sequence. The model’s failure to generalize, therefore, cannot be simply attributed to reversed input sequences. Instead, it reflects a deeper limitation in semantic comprehension and situational reasoning. In particular, the model appears to rely heavily on surface-level “data” memory of linguistic forms, while lacking deeper “knowledge” memory that supports flexible reasoning. This observation underscores the broader challenge in bridging memorization and true understanding: current LLMs may perform well in rote recall but struggle with functional language competence necessary for semantic integration and scenario-based reasoning. Enhancing scenario cognition may therefore be a key step toward bridging the gap between “data” memory and genuine “knowledge” memory.

Furthermore, a deeper analysis of the erroneous outputs revealed a potential correlation between the model’s situational cognition ability and its tendency to produce hallucinations. As shown in Figure~\ref{fig:case_study}, most errors do not display obvious grammatical or pragmatic flaws, yet they deviate substantially from factual correctness. Drawing on these observations, we argue that hallucination in LLMs is, at least in part, a reflection of their insufficient situational cognition. Specifically, the model’s “data” memory provides strong competence in formal linguistic patterns, allowing it to generate grammatically fluent and coherent text. However, the lack of robust “knowledge” memory limits its ability to verify the semantic accuracy of its outputs or to integrate factual information effectively, the model appears confined to producing content that seems plausible on the surface but fails to grasp the semantics of its outputs, ultimately leading to certain types of hallucinations.

\section{Related Work}
\subsection{Cognitive Capabilities of LLMs}
Research on LLMs’ cognitive abilities reveals strengths in language processing but persistent challenges in reasoning and functional competence \citep{webb2023emergent}. \citet{niu2024large} highlight LLMs’ human-like language processing yet note deficits in reasoning with novel prompts and context-dependent understanding. \citet{lamprinidis2023large} show LLMs’ high error rates in limited-data inductive reasoning, underperforming Bayesian predictors. \citet{mahowald2024dissociating} argue LLMs excel in formal linguistic competence but struggle with functional competence, lacking deep semantic understanding. \citet{binz2023using} find GPT-3 limited in causal reasoning and deliberation, indicating poor generalization beyond training data. \citet{ullman2023large} demonstrate LLMs’ failure in altered Theory-of-Mind tasks, suggesting weak cognitive modeling. \citet{blank2023what} emphasize methodological pitfalls in assessing LLMs’ cognitive capacities, advocating for rigorous language-based evaluations. Recently, \citet{zhao2025explaining} also highlight the importance of implicit cognitive knowledge beyond the given text and propose a post-hoc knowledge probing approach to explain and evaluate the cognitive abilities of black-box LMs after training, thereby bridging them with human-understandable cognition. These findings align with our study, which explores LLMs’ scenario cognition, revealing their reliance on surface-level ``data'' memory hinders semantic integration of multiple scenario elements, underscoring gaps in human-like knowledge memory.

\subsection{Knowledge Memory in LLMs}
Research on LLMs' knowledge memory reveals significant limitations in generalizing learned associations. \citet{berglund2024the} demonstrate the ``Reversal Curse'', where LLMs trained on ``$A$ is $B$'' fail to infer ``$B$ is $A$'', suggesting reliance on surface-level ``data'' memory over deeper ``knowledge'' memory. Similarly, \citet{grosse2023studying} use influence functions to show that training examples matching the input order dominate LLM outputs, with reverse-order examples having minimal impact. \citet{meng2022locating} further indicate that factual associations are stored directionally in LLMs, complicating bidirectional recall. \citet{petroni2019language} explore LLMs as knowledge bases, noting their struggle with consistent factual retrieval under varying prompts. Additionally, \citet{elazar2021measuring} highlight inconsistencies in LLM outputs, attributing them to a lack of robust semantic understanding. Unlike these studies, which focus on simple relations or factual recall, our work investigates scenario cognition in complex, multi-role contexts, evaluating both model outputs and internal representations to underscore persistent gaps in semantic integration and knowledge memory.

\section{Conclusion}
This study is the first to assess the scenario cognition capabilities of LLMs by introducing a bi-perspective evaluation framework from both the output and internal representation perspectives with a scenario-based dataset. Our findings indicate that, although LLMs are capable of accurately memorizing Atomic Knowledge from the Memory Set, they struggle to answer questions involving specific scenario elements and fail to effectively encode the associations between scenario elements and arguments within their internal representations. These results suggest that \textbf{current LLMs do NOT have the ability of scenario cognition} and rely primarily on surface-level memorization rather than true semantic understanding or meaningful knowledge retention. Moreover, a brief case study reveals a potential link between limited scenario cognition and the occurrence of hallucinations in LLMs, offering a cognitive perspective that may inform future directions for improving model design and training.

\section*{Limitations}

While this study provides insights into the scenario cognition ability of LLMs, several limitations remain. First, our evaluation is based on a synthetic dataset of fictional facts, which may not fully capture the complexity and variability of real-world language scenarios. Second, the dataset scale is relatively small, potentially limiting the generalizability of the findings. Third, the probing analysis focuses on simple associations between scenario elements and arguments, which may overlook more complex or distributed semantic representations. Future work could address these limitations by expanding the dataset, incorporating real-world data, and exploring more advanced probing methods to provide deeper insights into the cognitive capabilities of LLMs.

\section*{Acknowledgments}
This work was supported by the Science and Technology Cooperation and Exchange Special Project of Shanxi Province (No.202204041101016), the National Natural Science Foundation of China (No.62376144), the Key Research and Development Program of Shanxi Province (No.202102020\\101008), and the Natural Language Processing Innovation Team (Sanjin Talents) Project of Shanxi Province.

\appendix
\section{Computational Resources}
To conduct the experiments described in this paper, we utilized NVIDIA V100 GPUs (32 GB) for training LLMs and all experiments were performed on a cluster of 4 $\times$ V100 GPUs.

Table~\ref{tab:flops} presents the estimated floating-point operations (FLOPs) consumed during the supervised fine-tuning of each model on the scenario-based dataset. FLOPs were calculated based on the model architecture, dataset size, and number of training epochs. These estimates provide insight into the computational cost of training each model family.

\begin{table}[h]
\centering
\begin{tabular}{lc}
\toprule
\textbf{Model} & \textbf{Training TFLOPs} \\
\midrule
Gemma2-2B & 1.526 \\
Gemma2-9B & 3.808 \\
\midrule
LLaMA3.2-1B & 0.934 \\
LLaMA3.2-3B & 1.514 \\
LLaMA3.1-8B & 2.737 \\
\midrule
Qwen2.5-0.5B & 0.574 \\
Qwen2.5-1.5B & 1.163 \\
Qwen2.5-3B & 1.939 \\
Qwen2.5-7B & 2.675 \\
Qwen2.5-14B & 4.369 \\
\bottomrule
\end{tabular}
\caption{Computational cost (in TFLOPs) for training each model on the scenario-based dataset.}
\label{tab:flops}
\end{table}

\section{Probing Design Discussion}
\label{sec:appendix-probe}
In Section~\ref{sec:Perspective of Internal Representations}, we introduced a linear probe to assess whether LLMs encode associations between scenario elements and their arguments in internal representations. Here, we explore three alternative probing designs: SimilarityMLP, EnhancedSimilarityMLP, and an Attention-based probing approach. These designs aim to capture potentially non-linear or complex interactions that a simple linear probe might miss, serving as extensions to evaluate internal encodings from different angles. We describe each method and summarize their experimental results, which further support the main findings that LLMs lack robust internal encoding of scenario cognition.

\subsection{SimilarityMLP}
The SimilarityMLP is a two-layer multilayer perceptron (MLP) designed to capture non-linear relationships, aiming to introduce non-linear transformations for richer feature representation. For a scenario element representation $\mathbf{h}_{e_i}^{\mathcal{L}} \in \mathbb{R}^d$ and argument representation $\mathbf{h}_{a_j}^{\mathcal{L}} \in \mathbb{R}^d$ at layer level $\mathcal{L}$, we concatenate them as:
\begin{equation}
\mathbf{z}_{i,j}^{\mathcal{L}} = [\mathbf{h}_{e_i}^{\mathcal{L}}; \mathbf{h}_{a_j}^{\mathcal{L}}] \in \mathbb{R}^{2d}
\end{equation}

The probe applies a non-linear transformation to predict whether $e_i$ and $a_j$ form a matching pair:
\begin{equation}
  \mathbf{h}_{i,j}^{\mathcal{L}} = \text{ReLU}(\mathbf{W}_1 \mathbf{z}_{i,j}^{\mathcal{L}} + \mathbf{b}_1)
\end{equation}
\begin{equation}
\hat{y}_{i,j}^{\mathcal{L}} = \sigma(\mathbf{W}_2 \mathbf{h}_{i,j}^{\mathcal{L}} + \mathbf{b}_2)
\end{equation}
where $\mathbf{W}_1 \in \mathbb{R}^{d \times 2d}$, $\mathbf{b}_1 \in \mathbb{R}^d$, $\mathbf{W}_2 \in \mathbb{R}^{2 \times d}$, $\mathbf{b}_2 \in \mathbb{R}^2$ are trainable parameters, and $\sigma(\cdot)$ is the sigmoid function yielding probabilities for binary classification.

\subsection{EnhancedSimilarityMLP}
The EnhancedSimilarityMLP extends SimilarityMLP by incorporating derived features to enhance sensitivity to representational differences. We compute the absolute difference $\mathbf{d}_{i,j}^{\mathcal{L}} = |\mathbf{h}_{e_i}^{\mathcal{L}} - \mathbf{h}_{a_j}^{\mathcal{L}}|$ and element-wise product $\mathbf{m}_{i,j}^{\mathcal{L}} = \mathbf{h}_{e_i}^{\mathcal{L}} \odot \mathbf{h}_{a_j}^{\mathcal{L}}$, forming the input:
\begin{equation}
  \mathbf{z}_{i,j}^{\mathcal{L}} = [\mathbf{h}_{e_i}^{\mathcal{L}}; \mathbf{h}_{a_j}^{\mathcal{L}}; \mathbf{d}_{i,j}^{\mathcal{L}}; \mathbf{m}_{i,j}^{\mathcal{L}}] \in \mathbb{R}^{4d}
\end{equation}

The probe applies a non-linear transformation:
\begin{equation}
\mathbf{h}_{i,j}^{\mathcal{L}} = \text{ReLU}(\mathbf{W}_1 \mathbf{z}_{i,j}^{\mathcal{L}} + \mathbf{b}_1)
\end{equation}
\begin{equation}
\hat{y}_{i,j}^{\mathcal{L}} = \sigma(\mathbf{W}_2 \mathbf{h}_{i,j}^{\mathcal{L}} + \mathbf{b}_2)
\end{equation}
where $\mathbf{W}_1 \in \mathbb{R}^{d \times 4d}$, $\mathbf{b}_1 \in \mathbb{R}^d$, $\mathbf{W}_2 \in \mathbb{R}^{2 \times d}$, $\mathbf{b}_2 \in \mathbb{R}^2$ are trainable parameters, and $\sigma(\cdot)$ is the sigmoid function.

\subsection{Attention-based Probing}
Motivated by the Attention mechanism in Transformer architectures prevalent in LLMs, we designed an Attention-based probe as a supplement to the MLP-based probes. It analyzes the Attention scores between target pairs (\textit{i.e.,} scenario elements and their corresponding argument pairs) and non-target pairs (\textit{i.e.,} scenario elements and unrelated tokens) to reveal potential interference in internal representations. For a scenario element representation $\mathbf{h}_{e_i}^{\mathcal{L}} \in \mathbb{R}^d$ and argument representation $\mathbf{h}_{a_j}^{\mathcal{L}} \in \mathbb{R}^d$ at layer level $\mathcal{L}$, we compute the Attention score as:
\begin{equation}
  \alpha_{i,j}^{\mathcal{L}} = \frac{\exp((\mathbf{W}_q \mathbf{h}_{e_i}^{\mathcal{L}})^\top (\mathbf{W}_k \mathbf{h}_{a_j}^{\mathcal{L}}) / \sqrt{d})}{\sum_{k} \exp((\mathbf{W}_q \mathbf{h}_{e_i}^{\mathcal{L}})^\top (\mathbf{W}_k \mathbf{h}_{a_k}^{\mathcal{L}}) / \sqrt{d})}
\end{equation}
where $\mathbf{W}_q, \mathbf{W}_k \in \mathbb{R}^{d \times d}$ are trainable weight matrices. Instead of binary classification, we analyze distributions such as average and maximum Attention scores for target vs. non-target pairs to assess relational encoding.

\subsection{Experimental Results}
\begin{figure*}[htbp]
  \includegraphics[width=\textwidth]{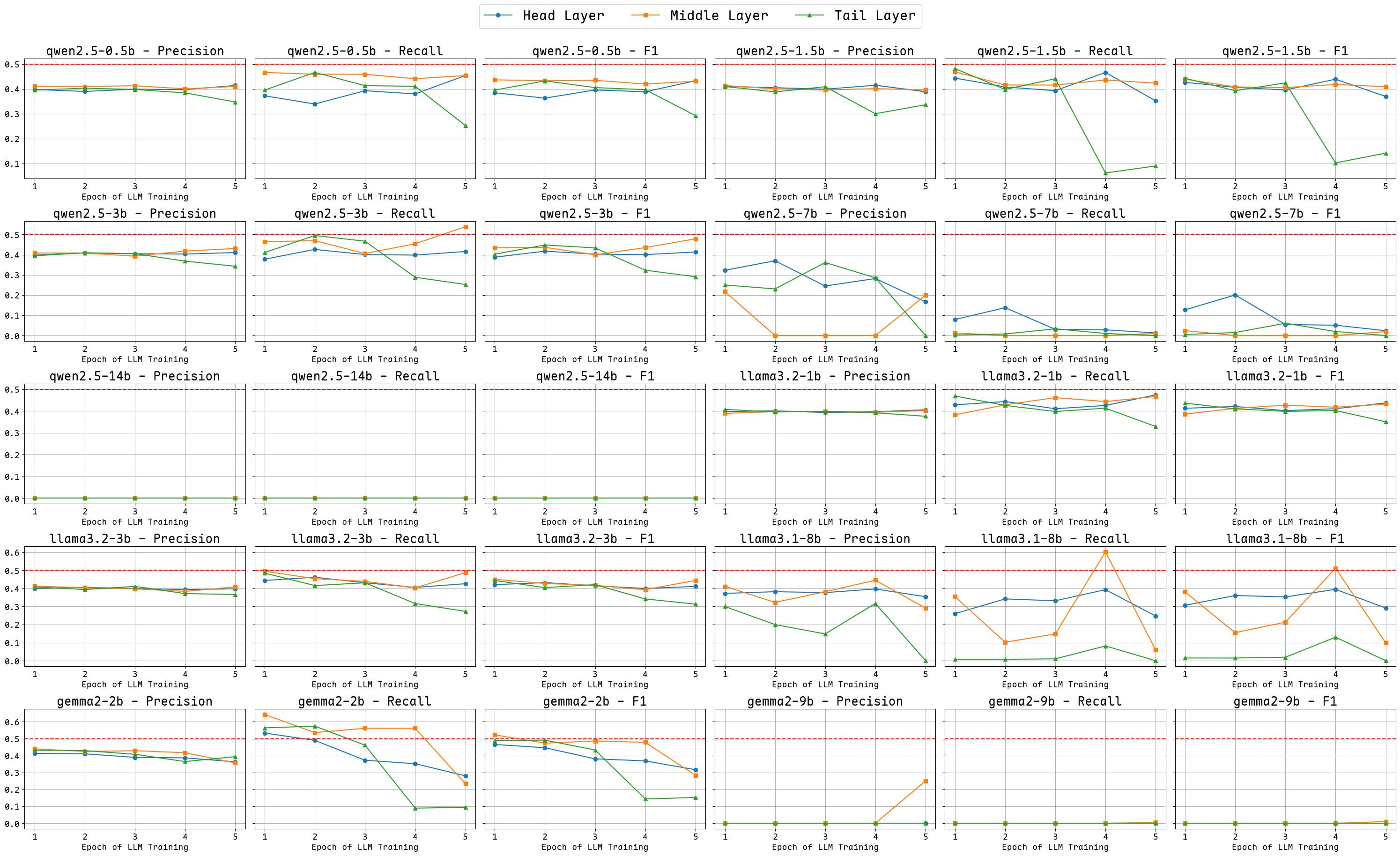}
  \caption{Visualization of probing results with \textbf{SimilarityMLP} for LLMs' internal representations across training epochs, with subfigures for Precision, Recall, and F1 scores at different Transformer layers (Head, Mid, Tail), where the red dashed line at 0.5 indicates the baseline for binary classification of scenario element-argument correspondences.}
  \label{fig:output_prob_mlp}
\end{figure*}

\begin{figure*}[htbp]
  \includegraphics[width=\textwidth]{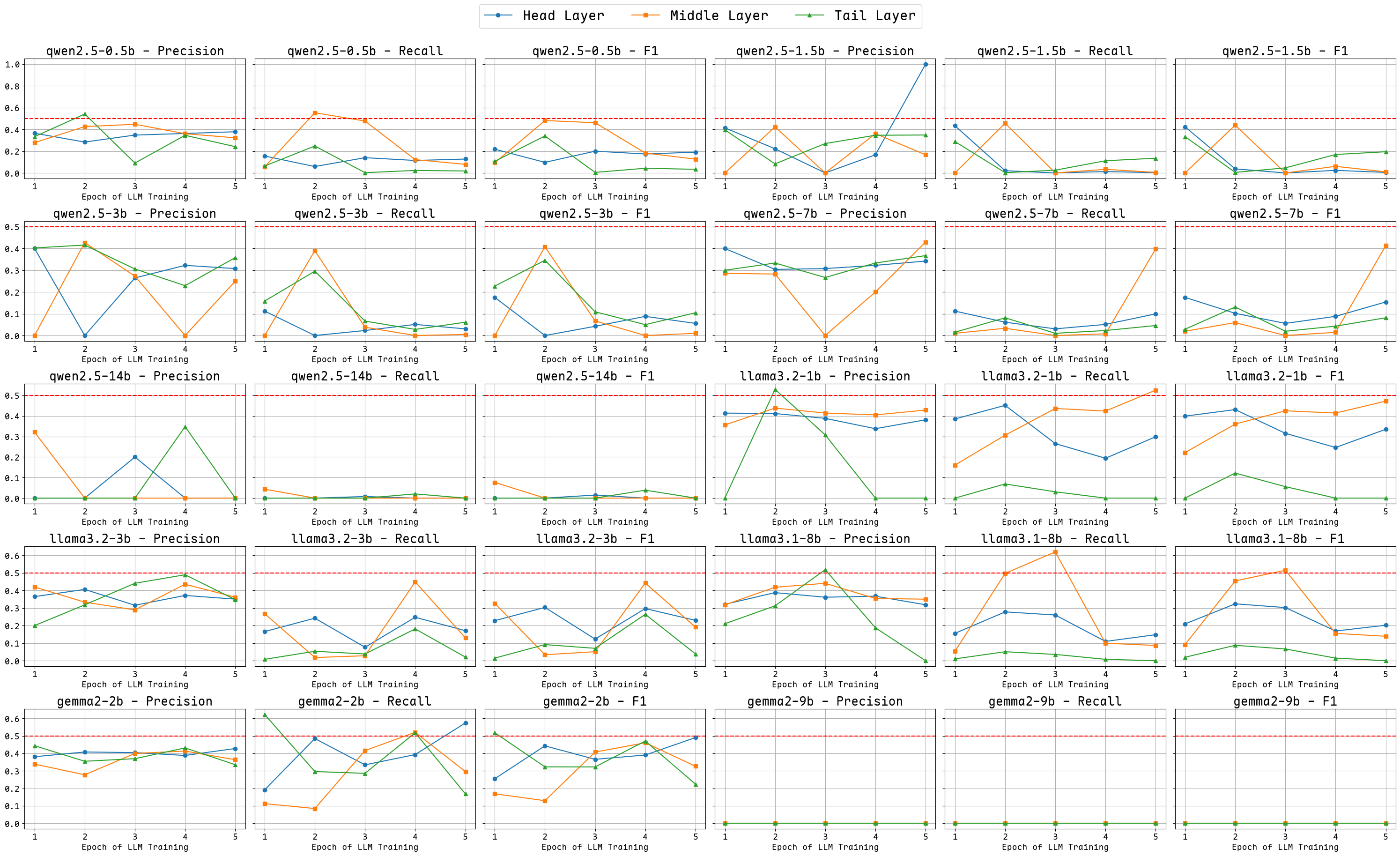}
  \caption{Visualization of probing results with \textbf{EnhancedSimilarityMLP} for LLMs' internal representations across training epochs, with subfigures for Precision, Recall, and F1 scores at different Transformer layers (Head, Mid, Tail), where the red dashed line at 0.5 indicates the baseline for binary classification of scenario element-argument correspondences.}
  \label{fig:output_prob_enhanced_mlp}
\end{figure*}

\begin{figure*}[htbp]
  \includegraphics[width=\textwidth]{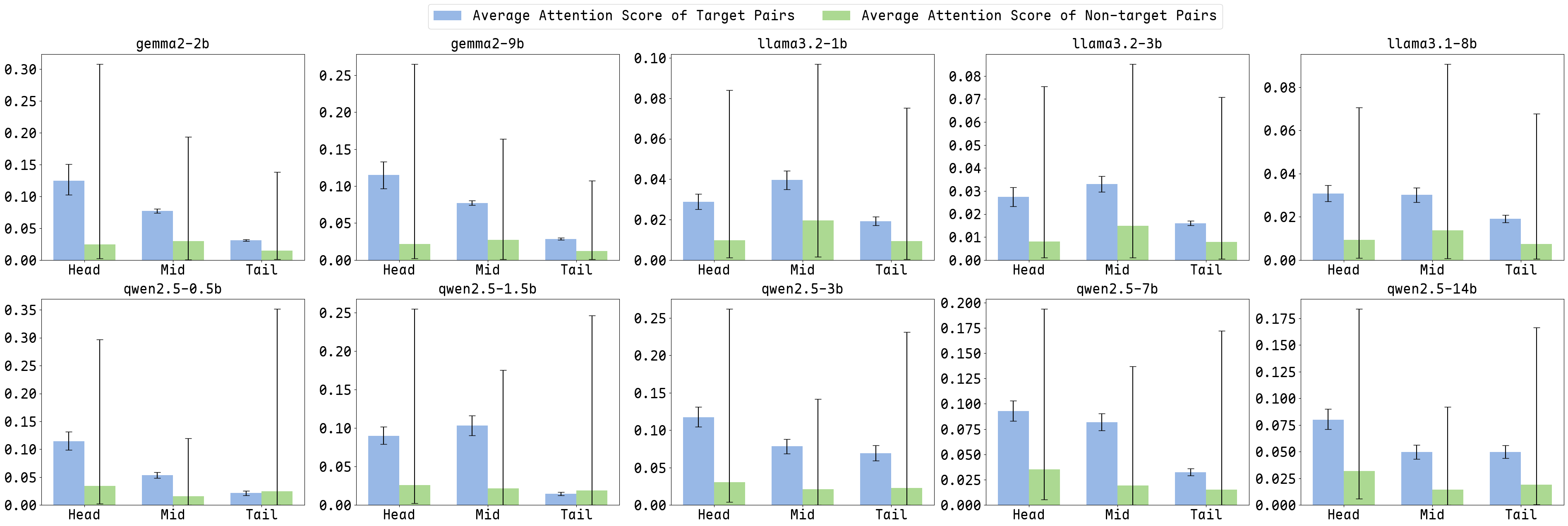}
  \caption{Visualization of probing results using \textbf{Attention-based Probing} for LLMs’ internal representations at the final epoch. Each subfigure shows the Attention scores of a specific LLM at different Transformer layers (Head, Mid, Tail), comparing target pairs (\textit{i.e.}, scenario elements and their corresponding arguments) with non-target pairs (\textit{i.e.}, scenario elements and unrelated tokens). Each bar represents the average value, and the error bar indicates the maximum and minimum values.}
  \label{fig:output_prob_att}
\end{figure*}

We evaluated the SimilarityMLP, EnhancedSimilarityMLP, and Attention-based probing method on the same set of LLMs as in the main analysis. The results are presented in Figures~\ref{fig:output_prob_mlp}, \ref{fig:output_prob_enhanced_mlp}, and \ref{fig:output_prob_att}, respectively.

As shown in Figures~\ref{fig:output_prob_mlp} and \ref{fig:output_prob_enhanced_mlp}, compared to the linear probe, both MLP-based methods exhibit relatively low performance across all metrics (Precision, Recall, and F1), with no clear correlations to model depth or training epochs. This indicates that introducing non-linear transformations and derived features does not significantly improve the detection of scenario element–argument associations, reinforcing the main conclusion that LLMs' scenario cognition is insufficient and disorganized. Notably, larger models like \textit{Qwen2.5-14B} and \textit{Gemma2-9B} often show near-zero scores, suggesting complex probes may overfit without uncovering deeper relational encodings.

As a supplement, the Attention-based probe (Figure~\ref{fig:output_prob_att}) reveals that LLMs partially learn Attention relationships for target pairs, with average Attention scores notably higher than for non-target pairs. However, the maximum Attention for non-target pairs remains significantly higher, indicating persistent interference from extraneous tokens outside the semantic scenario. This Attention-based analysis complements the MLP probes by highlighting how such interference may hinder clear scenario cognition, further underscoring the limitations in LLMs' internal representations.

\section{Prompt Templates for Data Generation}
\label{sec:appendix-templates}

\begin{figure}[p]
    \centering
    \includegraphics[width=\columnwidth]{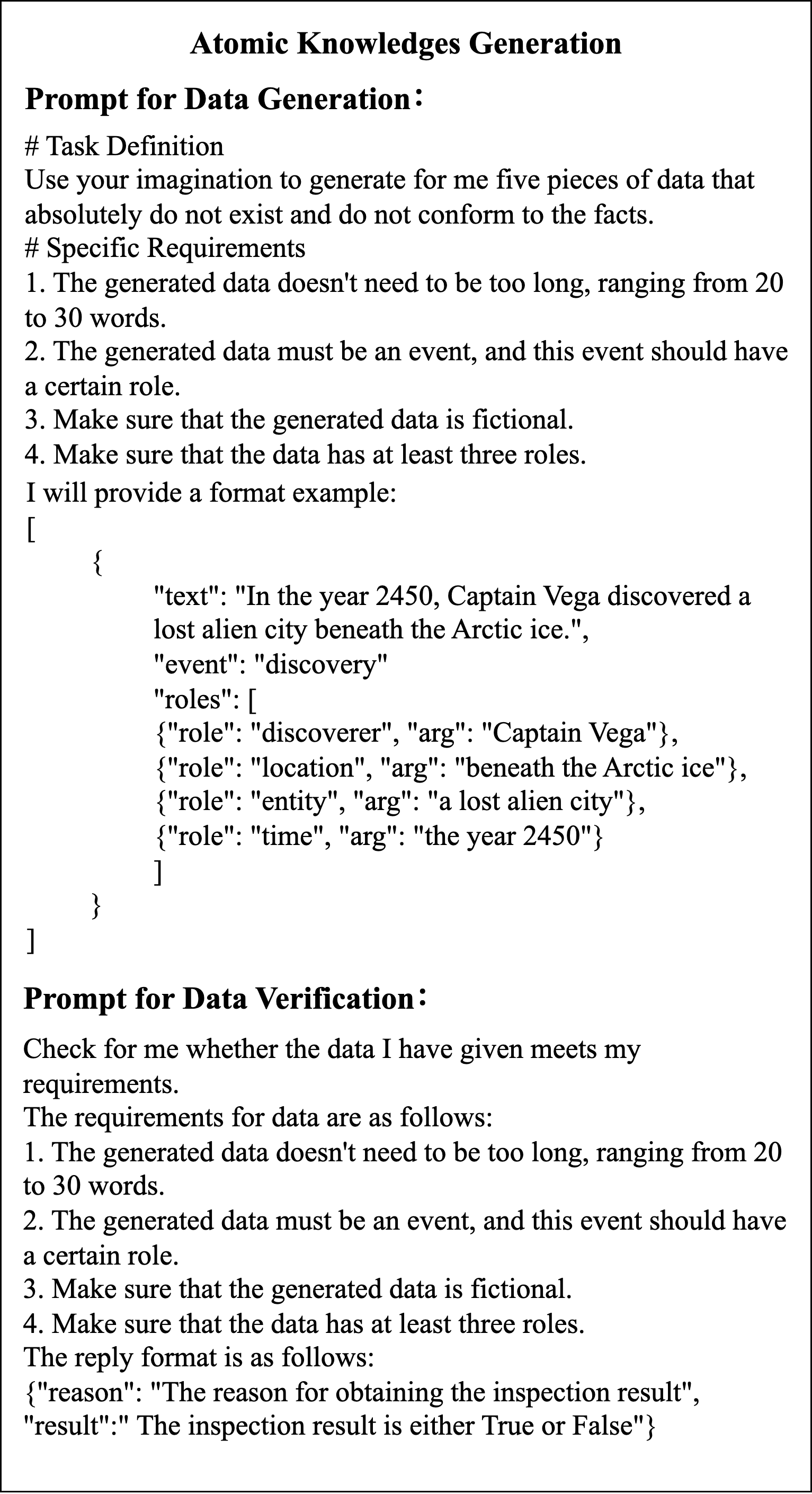}
    \caption{Prompt template for Atomic Knowledge Generation.}
    \label{fig:atomic_knowledge_prompt}
\end{figure}

\begin{figure}[p]
    \centering
    \includegraphics[width=\columnwidth]{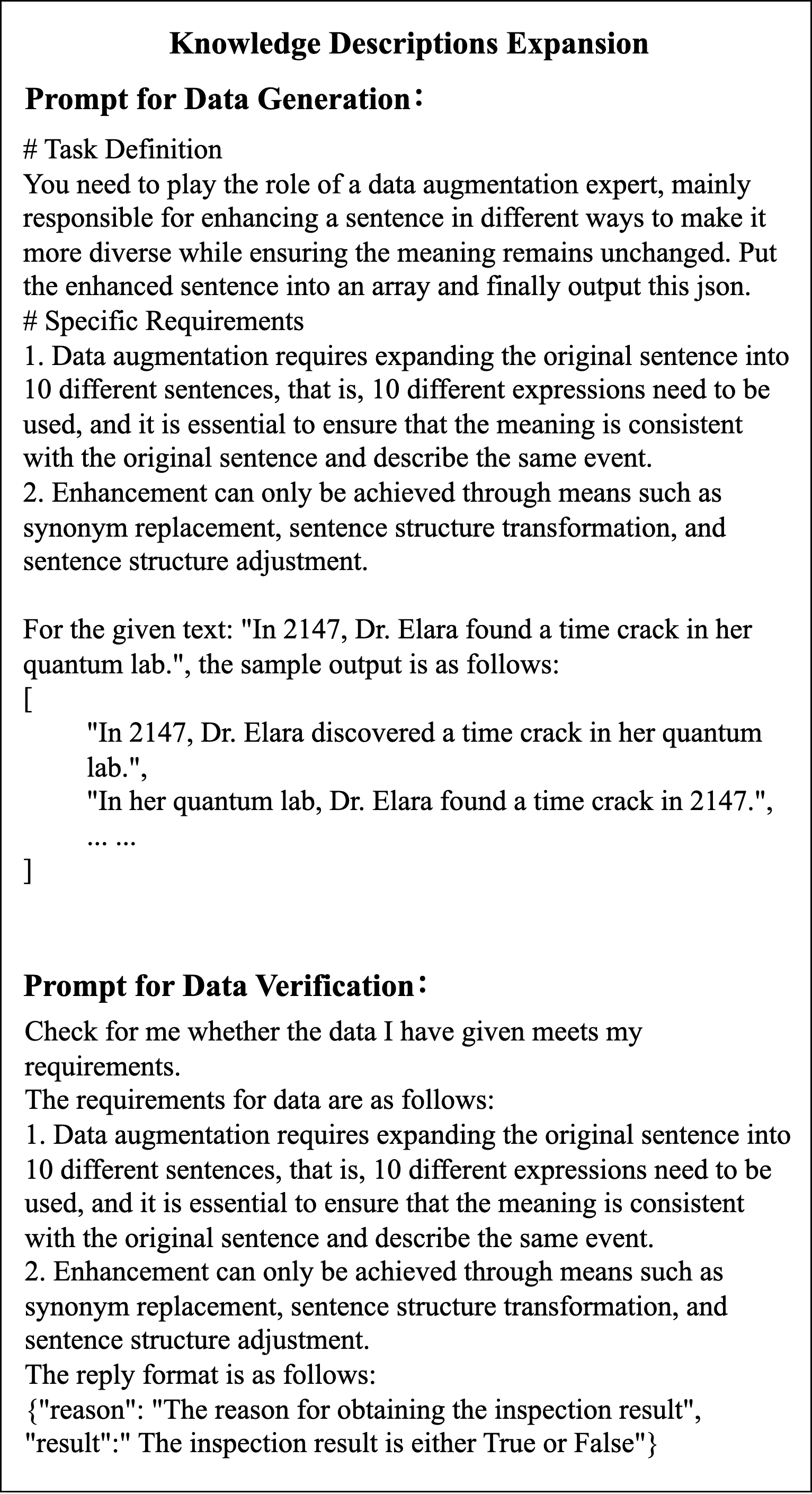}
    \caption{Prompt template for Knowledge Description Expansion.}
    \label{fig:description_expansion_prompt}
\end{figure}

This appendix provides details on the prompt templates used in the data generation and verification process for our scenario-based dataset, specifically in the stages of Section~\ref{sec:Atomic Knowledge Generation} Atomic Knowledge Generation (Figure~\ref{fig:atomic_knowledge_prompt}) and Section~\ref{sec:Knowledge Description Expansion} Knowledge Description Expansion (Figure~\ref{fig:description_expansion_prompt}). These templates were designed to guide LLMs in generating high-quality, diverse, and semantically consistent textual data to support the evaluation of scenario cognition.

\section{Examples of Generated Data}
\label{sec:appendix-example}
This appendix presents example datas of our scenario-based dataset. These examples illustrate the quality and characteristics of the generated data, which underpin the evaluation of LLMs scenario cognition capabilities. 

Table~\ref{tab:generated_datas} presents the examples which reflect the design principles of our training dataset, including fictionality, role richness, conciseness, and semantic consistency. For brevity, we only show three representative Expanded Descriptions per Atomic Knowledge but the full set of descriptions is available and have been used during our evaluation. These data support the evaluation of large language models' scenario cognition by providing diverse inputs for supervised fine-tuning, as discussed in Section~\ref{sec:Atomic Knowledge Generation} and \ref{sec:Knowledge Description Expansion}.

Table~\ref{tab:scenario_questions} presents examples of scenario-based questions generated which are designed to evaluate the scenario cognition capabilities of LLMs. Each question is derived from the corresponding Atomic Knowledge and its expanded descriptions, focusing on specific scenario elements and their relationships. The questions are structured to elicit responses that demonstrate the model's understanding of the scenario context and its ability to reason about the roles and actions involved as discussed in Section~\ref{sec:Scenario Element Annotation} and \ref{sec:Scenario Question Generation}.

\begin{table*}[h]
    \centering
    \resizebox{\textwidth}{!}{
    \begin{tabular}{p{0.45\textwidth} p{0.45\textwidth}}
        \toprule
        \textbf{Atomic Knowledge} & \textbf{Expanded Descriptions} \\
        \midrule
        Mathematician Dr. Lincoln Quantum presented a groundbreaking proof at the Annual Theoretical Mathematics Conference, confirming the existence of Quintilian particles. &
        \begin{itemize}
            \item At the Annual Theoretical Mathematics Conference, Mathematician Dr. Lincoln Quantum introduced a pioneering proof confirming the existence of Quintilian particles.
            \item A groundbreaking proof was presented by Mathematician Dr. Lincoln Quantum at the Annual Theoretical Mathematics Conference, which confirmed the existence of Quintilian particles.
            \item At the Annual Theoretical Mathematics Conference, Mathematician Dr. Lincoln Quantum offered new evidence through a groundbreaking proof of Quintilian particles' existence.
        \end{itemize} \\
        \midrule
        Captain James Morrison rescued endangered wildlife during a cruise in the Pacific Ocean. &
        \begin{itemize}
            \item During a cruise in the Pacific Ocean, Captain James Morrison rescued endangered wildlife.
            \item When Captain James Morrison was on a cruise in the Pacific Ocean, he rescued endangered wildlife.
            \item In the Pacific Ocean, Captain James Morrison's mission was to rescue endangered wildlife.
        \end{itemize} \\
        \midrule
        Pediatrician Lucy Armstrong successfully performed surgery to remove a mysterious malformation from a young patient. &
        \begin{itemize}
            \item During the surgery, Pediatrician Lucy Armstrong successfully removed a mysterious malformation from a young patient.
            \item To remove a mysterious malformation, Pediatrician Lucy Armstrong successfully carried out the surgery on a young patient.
            \item In successful surgery, Pediatrician Lucy Armstrong removed a mysterious malformation from a young patient.
        \end{itemize} \\
        \bottomrule
    \end{tabular}
    }
    \caption{Examples of Atomic Knowledge and their corresponding expanded descriptions, which compose the Memory Set.}
    \label{tab:generated_datas}
\end{table*}

\begin{table*}[h]
    \centering
    \resizebox{\textwidth}{!}{
    \begin{tabular}{p{0.45\textwidth} p{0.45\textwidth}}
        \toprule
        \textbf{Atomic Knowledge} & \textbf{Scenario-based Question} \\
        \midrule
        Mathematician Dr. Lincoln Quantum presented a groundbreaking proof at the Annual Theoretical Mathematics Conference, confirming the existence of Quintilian particles. &
        \begin{itemize}
            \item The discoverer who presented a groundbreaking proof confirming the existence of Quintilian particles at the Annual Theoretical Mathematics Conference is \underline{\textit{Mathematician Dr. Lincoln Quantum}}.
            \item The location where Mathematician Dr. Lincoln Quantum presented the groundbreaking proof confirming the existence of Quintilian particles is \underline{\textit{Annual Theoretical}} \underline{\textit{Mathematics Conference}}.
            \item The groundbreaking proof presented by Mathematician Dr. Lincoln Quantum at the Annual Theoretical Mathematics Conference confirmed the existence of \underline{\textit{Quintilian particles}}.
        \end{itemize} \\
        \midrule
        Captain James Morrison rescued endangered wildlife during a cruise in the Pacific Ocean. &
        \begin{itemize}
            \item The rescuer of endangered wildlife during a cruise in the Pacific Ocean is \underline{\textit{Captain James Morrison}}.
            \item Captain James Morrison rescued endangered wildlife during a cruise in the \underline{\textit{Pacific Ocean}}.
            \item The subject that Captain James Morrison was rescued during a cruise in the Pacific Ocean is \underline{\textit{endangered wildlife}}.
        \end{itemize} \\
        \midrule
        Pediatrician Lucy Armstrong successfully performed surgery to remove a mysterious malformation from a young patient. &
        \begin{itemize}
            \item The surgeon who successfully performed the surgery to remove a mysterious malformation from a young patient is \underline{\textit{Pediatrician Lucy Armstrong}}.
            \item The procedure that Pediatrician Lucy Armstrong successfully performed to remove a mysterious malformation from a young patient is \underline{\textit{surgery}}.
            \item The surgery performed by Pediatrician Lucy Armstrong successfully removed a mysterious malformation from \underline{\textit{a young patient}}.
        \end{itemize} \\
        \bottomrule
    \end{tabular}
    }
    \caption{Examples of scenario-based questions generated from Atomic Knowledge, which compose the Understanding Set. The underlined text indicates the expected answer.}
    \label{tab:scenario_questions}
\end{table*}

\end{document}